\documentclass[11pt,a4paper]{article}
\usepackage[hyperref]{acl2021}

\usepackage[T1]{fontenc}
\usepackage{bbold}
\usepackage{pifont}

\usepackage{times}
\usepackage{latexsym}
\usepackage{nert}

\usepackage[noend]{algpseudocode}
\usepackage[english]{babel}
\usepackage{microtype}
\aclfinalcopy 

\NewDocumentCommand\emojicandle{}{
    \includegraphics[scale=0.03]{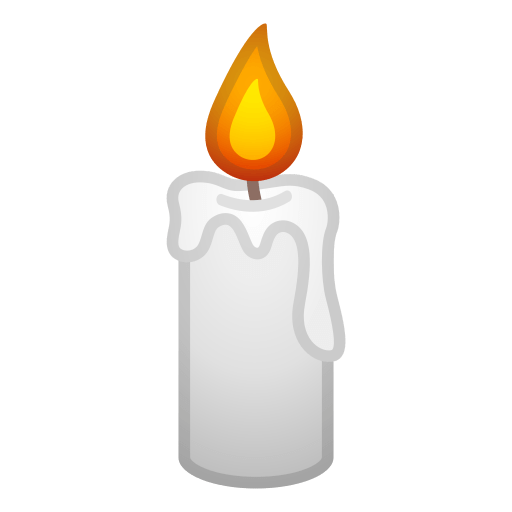}
}

\NewDocumentCommand\emojismallcandle{}{
    \includegraphics[scale=0.015]{./figures/candle-emoji-by-google.png}
}

\definecolor{bluegray_1}{rgb}{0.4, 0.6, 0.99}

\definecolor{ceruleanblue}{rgb}{0.16, 0.32, 0.75}

\definecolor{darkcoral}{rgb}{0.8, 0.36, 0.27}

\definecolor{airforceblue}{rgb}{0.36, 0.54, 0.66}

\newcolumntype{M}[1]{>{\centering\let\newline\\\arraybackslash\hspace{0pt}}m{#1}}
\newcolumntype{L}[1]{>{\centering\let\newline\\\arraybackslash\hspace{0pt}}m{#1}}
\newcolumntype{C}[1]{>{\centering\let\newline\\\arraybackslash\hspace{0pt}}m{#1}}
\newcolumntype{R}[1]{>{\centering\let\newline\\\arraybackslash\hspace{0pt}}m{#1}}
\newcolumntype{P}[1]{>{\centering\let\newline\\\arraybackslash\hspace{0pt}}m{#1}}
\newcolumntype{Q}[1]{>{\centering\let\newline\\\arraybackslash\hspace{0pt}}m{#1}}
\newcolumntype{S}[1]{>{\centering\let\newline\\\arraybackslash\hspace{0pt}}m{#1}}

\definecolor{caddback}{rgb}{0.90, 0.98, 0.96}
\definecolor{cadd}{rgb}{0, 0.47, 0.34}
\definecolor{cdelback}{rgb}{1, 0.94, 0.92}
\definecolor{cdel}{rgb}{0.83, 0.32, 0.16}
\definecolor{ao(english)}{rgb}{0.0, 0.5, 0.0}
\definecolor{auburn}{rgb}{0.43, 0.21, 0.1}
\definecolor{airforceblue1}{rgb}{0.36, 0.54, 0.99}
\definecolor{brandeisblue}{rgb}{0.0, 0.44, 0.9}
\definecolor{blush}{rgb}{0.87, 0.36, 0.51}
\definecolor{darklavender}{rgb}{0.45, 0.31, 0.59}

\newcommand\CND{\textcolor{cdel}}
\newcommand\CSQ{\textcolor{auburn}}
\newcommand\ALT{\textcolor{blush}}
\newcommand\FA{\textcolor{brandeisblue}}
\newcommand\SA{\textcolor{ao(english)}}
\newcommand\TA{\textcolor{darklavender}}

\newcommand*{\email}[1]{\texttt{#1}}

\title{\emojicandle CANDLE: Decomposing Conditional and Conjunctive Queries for Task-Oriented Dialogue Systems}

\author{
Aadesh Gupta, Kaustubh D. Dhole, Rahul Tarway, Swetha Prabhakar,\\
\textbf{Ashish Shrivastava}\\
Amelia Science, IPsoft R\&D \\
\email{\href{firstname.lastname@ipsoft.com}{firstname.lastname@ipsoft.com}}
}

\date{}

\begin{document}
\maketitle

\begin{abstract}
Domain specific dialogue systems generally determine user intents by relying on sentence-level classifiers which mainly focus on single action sentences. Such classifiers are not designed to effectively handle complex queries composed of conditional and sequential clauses that represent multiple actions. We attempt to decompose such queries into smaller single-action sub-queries that are reasonable for intent classifiers to understand in a dialogue pipeline. We release,\emojismallcandle CANDLE~\footnote{We release CANDLE at\\ ~\href{https://github.com/aadesh11/CANDLE}{https://github.com/aadesh11/CANDLE}}, 
 (Conditional \& AND type Expressions), a dataset consisting of 4282 utterances manually tagged with conditional and sequential labels and demonstrates this decomposition by training two baseline taggers.
\end{abstract}


\section{Introduction}

In task-oriented dialogue systems, understanding the nuances of natural language, popularly by classifying intents and entities, is a key step in deciphering users' intentions to perform subsequent dialogue actions. A popular and widely used component for such understanding has been the design of sentence-level intent classifiers~\cite{dhole2020resolving,aacl-2020-asia-pacific,  casanueva-etal-2020-efficient} because of the ease of annotating sentences as well as the stupendous progress of text classification~\cite{mehri2020dialoglue, devlin-etal-2019-bert}. However, complex user queries cover a wider plethora of natural language niceties involving multi-intent requests like conditional and compound statements. It is infeasible and often redundant to annotate all pairwise combinations of intents, especially when the individual intents are expected to trigger independent actions or processes.

We undertake a simple approach by performing a surface-level breakdown of such complex statements into simpler ones that can be processed individually. This delineation is advantageous in exploiting existing sentence-level intent classifiers without the need for further intent annotations.

\begin{figure}
\includegraphics[width=\linewidth]{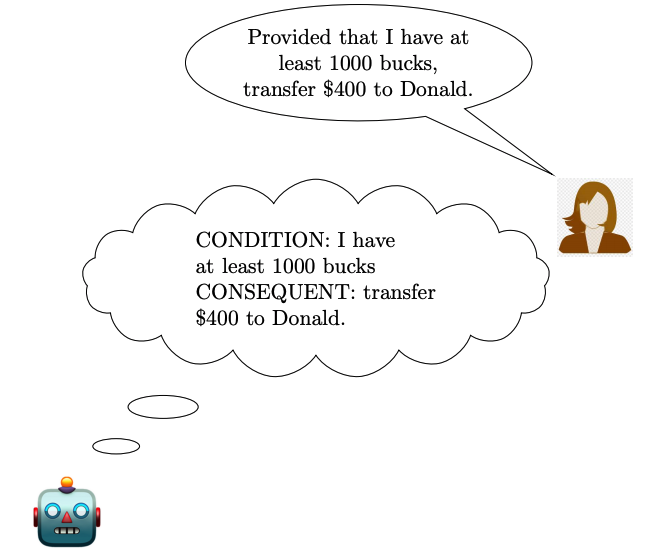}
\caption{A simple breakdown of a conditional user query.}
\label{fig:chat_comparison}
\end{figure}

In this work, we specifically focus on conditional and multi-intent statements as they are often encountered in dialogue. With conditionals, users can refer to situations whose existence is doubtful or whose outcomes are contingent on other events~\cite{rips1977suppositions}.

\ex.\label{ex:first} \emph{Provided that} \CND{$[I$ $have$ $at$ $least$ $1000$ $bucks$ $in$ $my$ $account]_\textbf{CND}$}, \CSQ{$[please$ $transfer$ $\$400$ $to$ $Donald]_\textbf{CSQ}$} \emph{otherwise} \ALT{[$check$ $my$ $account$ $balance]_\textbf{ALT}$}\footnote{Here, the annotations CND, CSQ, ALT, FA and SA refer to the Conditional, Consequence, Alternative, First action and Second action respectively.}


\ex.\label{ex:third} \FA{$[I$ $would$ $like$ $to$ $add$ $myself$ $to$ $the$ $insurance$ $policy]_\textbf{FA}$} \emph{and} \SA{$[my$ $wife's$ $bank$ $account]_\textbf{SA}$}.




For example, in case (1), it is imperative to first evaluate whether the user has \$1000 in his bank account. 
Cases like (2) can be split into multiple clauses - ``add myself to the insurance'' and ``add myself to my wife's account'' even possibly reconstructing elided predicates~\cite{schuster-etal-2018-sentences}.

Our analysis addresses two types of queries, firstly, conditional queries which require identifying the conditional and its paired consequent and an alternative, in rare cases. And second, multi-intent queries which require the breakdown into smaller intents to be executed sequentially.
The focus of this work is to decompose the individual actions of such statements. In Section 2, we first review previous approaches in handling conditional statements. In Section 3, we present \emojismallcandle CANDLE, a dataset spanning conditional and conjunctive annotations. Section 4 shows two baseline approaches used for tagging multiple intents. Finally, Section 5 describes our experiments and results.

\section{Related Work}
A fairly diverse amount of work has been devoted to the study and grammatical analysis of conditional statements in the past few decades~\cite{dudman1984conditional, chambers_1895, karttunen1973presuppositions} along with a lot of renewed interest recently~\cite{saha-mausam-2018-open, saha-etal-2020-conjnli, jeretic-etal-2020-natural}.~\citet{narayanan-etal-2009-sentiment} point out that conditional statements have unique characteristics that make it hard to determine the orientation of sentiments on topics and features.~\citet{Puente2009ExtractionOC, 5584115} emphasize that causality generally emerges from the entailment relationship between the antecedent and the consequent and use this idea to build causal chains for knowledge-based question answering.~\citet{TUW-223973} describe a pattern-based approach to identify the condition and the consequent.~\citet{alconditional} examined the patterns and frequency of conditional constructions in English and investigated the most frequently used subjunctive conditional collocates.
Besides, syntactic approaches have been quite popular to simplify clauses embedded with subordination and coordination.~\citet{posadas2017algorithm} extract syntactic n-grams by traversing over dependency tree.~\citet{dornescu-etal-2014-relative} investigate non-destructive simplification, a technique to extract embedded clauses from complex sentences using a tagging-based approach.~\citet{kim2021linguist} identify failed presuppositions within questions to improve question answering. The closest work to ours is~\citet{vo2021recognizing} which split sentences into conditional and resultant clauses, using rule-based and BERT~\cite{devlin-etal-2019-bert} based models. Our dataset is larger and covers diverse fine-grained annotations like ALTERNATIVE, SECOND\_ACTION, etc. 
To our knowledge, \emojismallcandle CANDLE is the first publicly released dataset.

\section{Dataset}

We first manually write a set of seed utterances with common conditional and conjunctive markers which span the domains of Banking, Telecom, Sports, Healthcare, and Information Technology. We try to increase the diversity of expressing conditionals by incorporating common and uncommon if-else constructions like~\textit{ if},~\textit{on the condition that},~\textit{provided that},~\textit{only if},~\textit{unless}, etc. We further augment this set by generating paraphrases by utilizing a T5 Transformer~\cite{wolf-etal-2020-transformers, JMLR:v21:20-074} fine-tuned on paraphrase pairs.~\footnote{We use the T5 model available at\\ ~\href{https://huggingface.co/ramsrigouthamg/t5_paraphraser}{https://huggingface.co/ramsrigouthamg/t5\_paraphraser}} Additionally, we also search for sentences of Wikipedia passages matching the above conditional patterns. We restrict our Wikipedia search to the paragraphs released by SQuAD~\cite{rajpurkar-etal-2016-squad}. We eventually perform 2-3 iterations of manual editing of these sentences to discount for paraphrasing errors. 

\paragraph{Annotation}
We conducted two rounds of annotation. We first requested annotators to tag segments of text that spanned one of the below tags. In the second round, the authors manually verified all the annotations.
\begin{itemize}
\item CONDITIONAL (\CND{CND}): The span of text describing a condition.
\item CONSEQUENCE (\CSQ{CSQ}): The span of text describing a consequence corresponding to a condition.
\item ALTERNATIVE (\ALT{ALT}): The span of text describing an alternative corresponding to a condition.
\item FIRST\_ACTION (\FA{FA}): The span of text describing the first action in a sentence.
\item SECOND\_ACTION (\SA{SA}): The span of text describing the second action followed by the first action.
\item THIRD\_ACTION (\TA{TA}): The span of text describing the third action followed by the second action.
\item NONE (NN): The span of text that does not include any conditional or sequential                                   action phrases.\\
\end{itemize}

This resulted in the final dataset, consisting of 4282 sentences which we refer to as \emojismallcandle CANDLE. Table~\ref{tab:tagsplits} describes the number of sentences corresponding to each label type. We randomly shuffle and create three splits - 3426 for training (80\%), 428 (10\%) for testing, and 428 (10\%) for validation. 

\begin{table*}[ht]
  \centering
  \begin{tabular}{c c c c}
  \hline
  \begin{tabular}{c}
    \textbf{Tag Type}
  \end{tabular} & \begin{tabular}{c}
  \textbf{\#tag(Training)}
  \end{tabular} & \begin{tabular}{c}
  \textbf{\#tag(Validation)}
  \end{tabular} & \textbf{\#tag(test)} \\
  \hline
     ALTERNATIVE&795&108&100\\
     CONDITIONAL&2585&330&315\\
     CONSEQUENCE&2584&330&315\\
     FIRST\_ACTION&645&77&78\\
     SECOND\_ACTION&645&77&78\\
     THIRD\_ACTION&165&20&24\\
     NONE&199&23&35\\
     \hline
  \end{tabular}
  \caption{Tag wise splits.}
  \label{tab:tagsplits}
\end{table*}

\begin{algorithm}[ht]
\SetAlgoLined
\KwData{user utterance}
\KwResult{tags}
\textbf{procedure} \textbf{\textit{RunEnsemble}}{$(utterance)$}\\
$parsed\gets depParse(utterance)$\\
$expand\_parsed\gets expandClauses(parsed)$\\
$grammar\_based\gets$ \textbf{\textit{GRM}}(expand\_parsed)\\
\eIf{grammar\_based $!=$ null}
{\Return $grammar\_based$
    }{
      \Return \textbf{\textit{ModelPredict}}(expand\_parsed)
    }
\caption{Ensemble Processing}
\label{alg:ensemble}
\end{algorithm}

\section{Baseline Approaches}
To set initial performance levels on \emojicandle CANDLE, we present two baseline approaches, an ensemble approach and an end-to-end transformer model. The Ensemble approach combines a set of pattern matching rules, a set of dependency rules, and a BiLSTM model with a CRF layer. In addition to that, we fine-tune a RoBERTa large~\cite{Liu2019RoBERTaAR} model.

\subsection{Ensemble Approach}
As shown in the Algorithm~\ref{alg:ensemble}, we first write rules over dependency fragments of the sentence to perform clausal expansion to reconstruct the sentence. Then we check if this sentence matches any of the grammar rules. 
If it doesn't we query the BiLSTM+CRF tagger trained on \emojicandle CANDLE.\\
\\
Below are the sub-tasks of our ensemble model:
\begin{enumerate}
{\bf \item Sentence Restructuring:}\\
At inference time, we first try to restructure a complex sentence and expand the clauses contained in it to a structure that is easier for the tagger. We detect clausal conjunctions using a dependency rule. If there is a conjunction present (i.e. the existence of a CONJ dependency relation) and a parent headed by XCOMP or CCOMP or ROOT, we use it to expand the second action. This is a lighter form of gapping~\cite{schuster2018sentences}. The below snippet shows examples being restructured:

\begin{itemize}
\item \textit{Transfer \$400 to John and Sam. --> Transfer \$400 to John and \emph{\colorbox{cyan!40}{Transfer \$400 to}} Sam.}\\

 \end{itemize}

We found that expanding the clauses in this fashion generally gives better results for tagging. Additionally, individual clauses tend to approximate their original single-action counterparts.
\\
{\bf \item Grammar Rule-Based Phrase Extraction:}
We define a set of ordered grammar rules as illustrated in Figure~\ref{fig:grammar_rule}. These rules (shown in Algorithm~\ref{alg:grammar} in Appendix) perform classic regex expression matching with each of the patterns. 

They are highly precise in recognizing conditional phrases. Although they provide excellent precision, their scope is limited and they lack the ability to handle complex statements, making them extremely restrictive.\\

{\bf \item Sequence-to-Sequence model:} 

We further use a sequence-to-sequence model as our second baseline approach. The model consists of a (Bidirectional LSTM)~\cite{GRAVES2005602} layer followed by a CRF~\cite{laffertyCrf} layer for tagging. If none of the grammar rules matches, the sentence is sent to this layer (Algorithm~\ref{alg:model} in Appendix). With the sequence-to-sequence model, we are able to increase the coverage as well as improve the tagging performance as shown in  Table~\ref{tab:results-text}. The BiLSTM-CRF model was trained on the training set of\emojicandle CANDLE for 25 epochs with a learning rate of 0.1.
\end{enumerate}

\begin{figure}
\includegraphics[width=\linewidth]{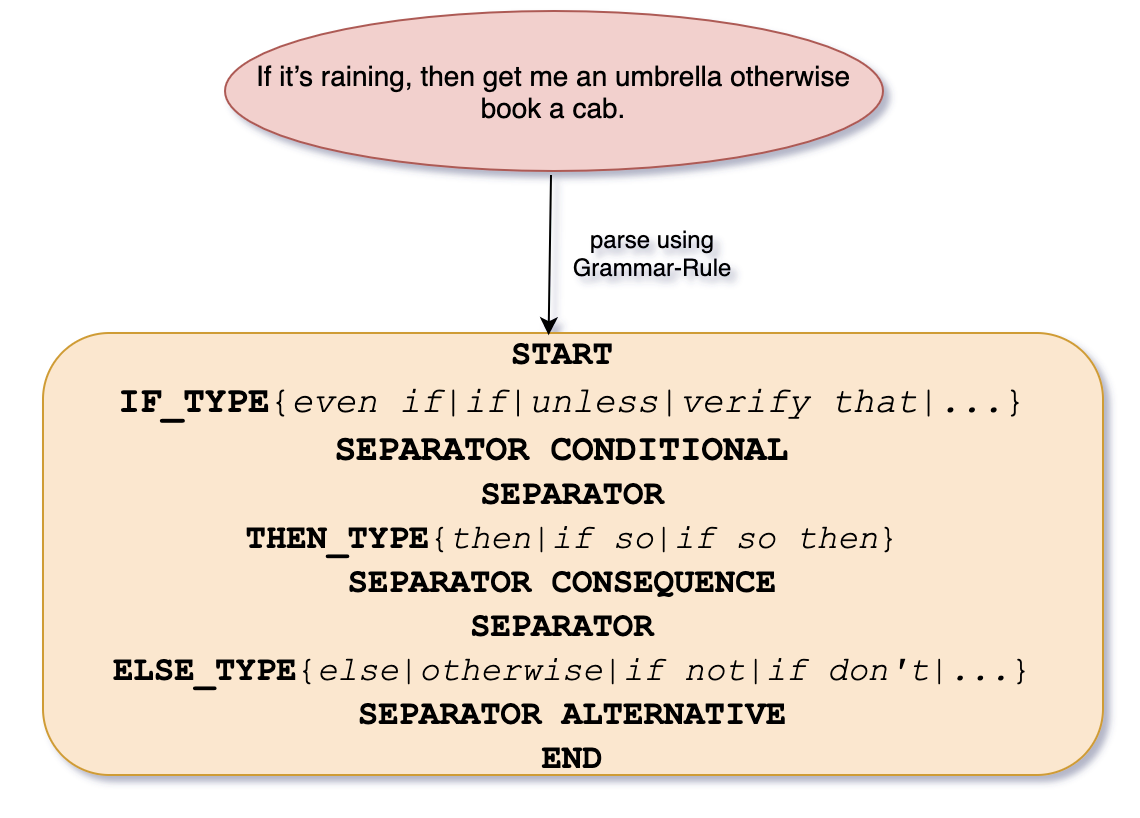}
\caption{Mapping sentence to Grammar Rule.}
\label{fig:grammar_rule}
\end{figure}

\begin{table*}[ht!]
  \centering
  \resizebox{\textwidth}{!}{
  \begin{tabular}{|l||c@{~~}cc|c@{~~}cc|c@{~~}cc|c@{~~}cc|c@{~~}cc|c@{~~}cc|c|c|}
    \hline
    \multirow{2}{*}{\textbf{Algorithm}} & \multicolumn{3}{c|}{\textbf{CND}} & \multicolumn{3}{c|}{\textbf{CSQ}} & \multicolumn{3}{c|}{\textbf{ALT}} & \multicolumn{3}{c|}{\textbf{FA}} & \multicolumn{3}{c|}{\textbf{SA}} & \multicolumn{3}{c|}{\textbf{TA}}&\\[0.15cm]
    \cline{2-19} & P & R & F1 & P & R & F1 & P & R & F1 & P & R & F1 & P & R & F1 & P & R & F1 & \textbf{Average} \\[0.2cm]
    \hline
    Rule-Based &89.40&63.08&73.97&82.98&54.42&65.73&92.00&49.46&64.34&100.00&03.90&07.50&100.00&05.19&09.88&-&-&-&55.60
    \\[0.2cm]
    BiLSTM &96.00&89.72&92.75&89.42&78.60&83.66&97.44&81.72&88.89&93.44&74.03&82.61&88.52&70.13&\textbf{78.26}&100.00&39.13&\textbf{56.25}&82.29\\[0.2cm]
    Ensemble (BiLSTM+Rule) &95.10&90.65&\textbf{92.82}&89.12&80.00&\textbf{84.31}&96.25&82.80&\textbf{89.02}&96.72&76.62&\textbf{85.51}&86.89&68.83&76.81&100.00&39.13&\textbf{56.25}&\textbf{82.75}\\[0.2cm]
    \hline
    BERT-base (multilingual-uncased) &95.57&95.87&95.72&95.22&94.92&95.07&99.01&100.00&\textbf{99.50}&88.46&88.46&88.46&87.01&85.90&86.45&91.30&87.50&89.36&93.92\\[0.2cm]
    XLM-R-base 
    &94.32&94.92&94.62&94.94&95.24&\textbf{95.09}&97.06&99.00&98.02&87.34&88.46&87.90&88.31&87.18&87.74&91.30&87.50&89.36&93.55\\[0.2cm]
    ELECTRA-base (discriminator) &95.56&95.56&95.56&94.03&94.92&94.47&99.01&100.00&\textbf{99.50}&90.54&85.90&88.16&88.31&87.18&87.74&91.67&91.67&91.67&93.56\\[0.2cm]
    \textbf{RoBERTa-base} 
    &94.32&94.92&94.62&94.03&94.92&94.47&97.06&99.00&98.02&95.95&91.03&\textbf{93.42}&89.61&88.46&\textbf{89.03}&92.31&100.00&\textbf{96.00}&\textbf{93.99}\\[0.2cm]
    \hline
    BERT-large (uncased)
    &94.95&95.56&95.25&94.01&94.60&94.30&98.02&99.00&98.51&90.79&88.46&89.61&87.84&83.33&85.53&95.24&83.33&88.89&92.83\\[0.2cm]
    XLM-R-large
    &95.86&95.56&\textbf{95.71}&94.03&94.92&94.47&99.01&100.00&\textbf{99.50}&93.67&94.87&\textbf{94.27}&87.18&87.18&87.18&91.67&91.67&\textbf{91.67}&94.37\\[0.2cm]
    \textbf{RoBERTa-large} 
    &95.86&95.56&\textbf{95.71}&95.86&95.56&\textbf{95.71}&97.06&99.00&98.02&94.81&93.59&94.19&89.47&87.18&\textbf{88.31}&87.50&87.50&87.50&\textbf{94.52}\\[0.2cm]
    \hline
  \end{tabular}
  }
  \caption{Sentence Splitting performance of different models on test set. P = Precision, R = Recall, CND = Conditional, CSQ = Consequence, ALT = Alternative, FA = First Action, SA = Second Action, TA = Third Action}
  \label{tab:results-text}
\end{table*}

\subsection{End-to-End Model}
Our end-to-end approach consists of a fine-tuned RoBERTa-large~\cite{Liu2019RoBERTaAR} model for the tagging task. RoBERTa is a retraining of BERT with improved training methodology, with more data and compute power. We fine-tune RoBERTa-large model over \emojicandle CANDLE using the BIO annotation format~\cite{ramshaw-marcus-1995-text}, for 5 epochs with a learning rate $5e-05$. We use a maximum sequence length of $128$, a batch size of $8$, a dropout rate of $0.1$. 

 


\section{Experiments and Results}

We performed our experiments with different setups as shown in Table~\ref{tab:results-text}. We clearly see that almost for all types of tags, the RoBERTa-large model outperformed the ensemble approach by a large margin and obtained the best F1 score.
We have not included the THIRD\_ACTION tag in our grammar rules, since it's difficult to handle longer utterances with grammar and also adds additional possibilities of over-matching.

Using only the BiLSTM-CRF model gave better results than the grammar rules: an F1 score of \textbf{82.29}. An ensemble approach of grammar and the BiLSTM-CRF gave us an F1 score \textbf{82.75}, a marginal improvement of \textbf{0.5\%}. Our final approach was the RoBERTa-large model that gave the best result in detecting all the clauses with F1 Score \textbf{94.52} - a significant \textbf{11.77\%} increase over our ensemble approach.

We also experimented with the Multilingual BERT-base~\cite{devlin2019bert}, XLM-R~\cite{Conneau2020UnsupervisedCR} and ELECTRA-base~\cite{Clark2020ELECTRAPT} models as mentioned in Table 2. 


Below are a few predictions from our best model which would be hard to cover with grammar rules.\\

\begin{enumerate}
\item Kindly verify that I have \$5000 in my saving, if yes, then move 3000 to checking, else apply for a loan. ---> Kindly verify that \CND{$[I$ $have$ $\$5000$ $in$ $my$ $saving]_\textbf{CND}$}, if yes, then \CSQ{$[move$ $3000$ $to$ $checking]_\textbf{CSQ}$}, else \ALT{[$apply$ $for$ $a$ $loan]_\textbf{ALT}$}.\\

\item As long as it's not raining, don't buy me a raincoat. It's cold outside. ---> As long as \CND{[$it's$ $not$ $raining]_\textbf{CND}$}, \CSQ{[$don't$ $buy$ $me$ $a$ $raincoat]_\textbf{CSQ}$}. It's cold outside.

\end{enumerate}
\section{Conclusion and Future Work}
In this study, we present \emojicandle CANDLE, a dataset labelled with conditional and conjunctive annotations to encourage the development of robust conditional systems. We further demonstrate few baseline approaches that decompose the sentence by tagging conditionals and conjunctive actions. We hope that task-oriented dialogue systems can take advantage of \emojicandle CANDLE to better understand complex user queries spanning multiple intents. In the future, it may be worthwhile to annotate datasets with overlapping spans.

\bibliographystyle{acl_natbib}
\bibliography{anthology,acl2021}

\clearpage
\appendix

\section{Appendix}
\label{sec:appendix}

\subsection{Supporting Algorithms}

\begin{algorithm}[ht]
\SetAlgoLined
\KwData{expanded-parsed user utterance}
\KwResult{mapped graph}
\textbf{procedure} \textbf{\textit{GRM}}{$(expand\_parsed)$}\\
$utterance\gets$ expand\_parsed as text\\
$grammar\_rules\gets$predefined grammar rules\\
\For {each rule in grammar\_rules} {
    \If{rule matched the utterance} {
        extract annotations from matched template\\
        $annotations\gets$ extracted annotation\\
        \Return \textbf{\textit{CreateGraph}}(annotations)\\
    }
}
\caption{Grammar-Rule Matching Algorithm (GRM)}
\label{alg:grammar}
\end{algorithm}

\begin{algorithm}[ht]
\SetAlgoLined
\KwData{expanded-parsed user utterance}
\KwResult{mapped graph}
\textbf{procedure} \textbf{\textit{ModelPredict}}{(expand\_parsed)}\\
$node\_set\gets$extractDNNModel(expand\_parsed)\\
\For{each annotation in node\_set} {
    $annotations\gets$ $extract$ $sentence$ $associated$ $with$ $tags$
    }
    $mapped\_graph\gets$ \textbf{\textit{CreateGraph}}(annotations)\\
    \Return \textit{mapped\_graph}
\caption{Model based Algorithm}
\label{alg:model}
\end{algorithm}

\begin{algorithm}[!ht]
\SetAlgoLined
\KwData{extracted annotations}
\KwResult{graph}
\textbf{procedure} \textbf{\textit{CreateGraph}}{$(annotations)$}\\
\For{each graph in graph template}{
    map $annotations$ with graph nodes\\
    \If{all nodes finds a map} {
        \Return graph
    }
}
\Return empty
\caption{Create graph algorithm}
\label{alg:graph}
\end{algorithm}

\begin{figure}
\includegraphics[width=\linewidth]{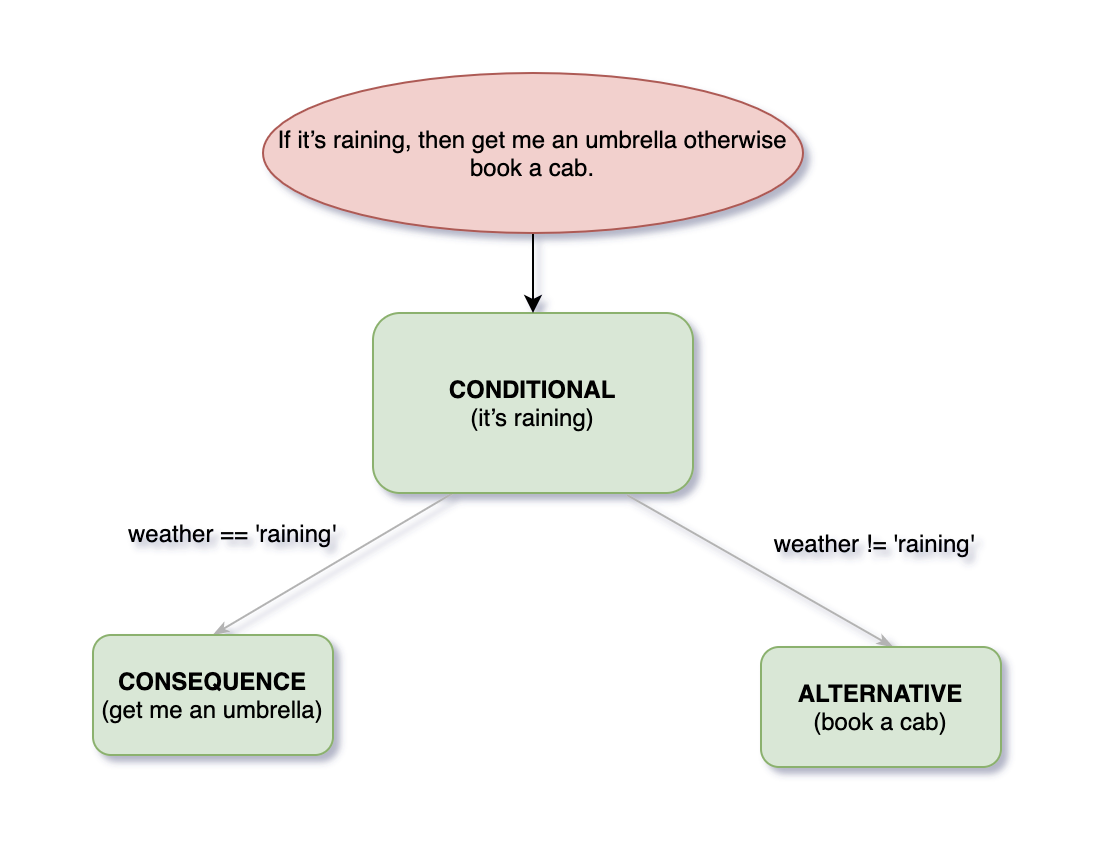}
\caption{Conditional graph template}
\label{fig:conditional_graph}
\end{figure}

\subsection{Mapping to Business Logic}
While, the way these extracted conditional tags map to real-life business processes is beyond the scope of this paper, we mention it briefly for the benefit of interested readers.
A post-processing step (Algorithm~\ref{alg:graph}) is used to convert the extracted tags into a readable graph, finally mapping it to domain-specific business logic. For this, we define a specific graphical structure to map these tags as shown in Figure~\ref{fig:conditional_graph}.\\
This allows a complex sentence to be mapped into a chain of business processes that can ultimately be executed on meeting a condition or a fixed criteria.

\end{document}